\documentclass[runningheads]{llncs}
\usepackage{graphicx}
\usepackage{subcaption}
\usepackage{amssymb}
\usepackage{amsmath}
\usepackage{xcolor}
\usepackage{dsfont}
\usepackage[utf8]{inputenc}
\usepackage{booktabs,siunitx,amsmath,caption, subcaption}
\usepackage{hyperref}

\begin{document}
\title{fMRI Neurofeedback Learning Patterns are Predictive of Personal and Clinical Traits}
\titlerunning{Predictive fMRI Learning Patterns}
% If the paper title is too long for the running head, you can set
% an abbreviated paper title here
%
\author{Rotem Leibovitz\inst{1} \and
Jhonathan Osin\inst{1} \and
Lior Wolf\inst{1}\and\\
Guy Gurevitch\inst{2,4}\and
Talma Hendler\inst{2,3,4,5}}

% index{Leibovitz, Rotem}
% index{Osin, Jhonathan}
% index{Wolf, Lior}
% index{Gurevitch, Guy}
% index{Hendler, Talma}

%
\authorrunning{R. Leibovitz, J. Osin, L. Wolf et al.}
% First names are abbreviated in the running head.
% If there are more than two authors, 'et al.' is used.
%
\institute{School of Computer Science, Tel Aviv University \and
Sagol Brain Institue, Tel-Aviv Sourasky Medical Center \and
% \email{lncs@springer.com}\\
% \url{http://www.springer.com/gp/computer-science/lncs} \and
School of Psychological Sciences, Tel Aviv University \and
Sackler Faculty of Medicine, Tel Aviv University \and
Sagol School of Neuroscience, Tel Aviv University}
% %
\maketitle              % typeset the header of the contribution

\setcounter{footnote}{0}    

\begin{abstract}
  We obtain a personal signature of a person's learning progress in a self-neuromodulation task, guided by functional MRI (fMRI). The signature is based on predicting the activity of the Amygdala in a second neurofeedback session, given a similar fMRI-derived brain state in the first session. The prediction is made by a deep neural network, which is trained on the entire training cohort of patients. This signal, which is indicative of a person's progress in performing the task of Amygdala modulation, is aggregated across multiple prototypical brain states and then classified by a linear classifier to various personal and clinical indications. The predictive power of the obtained signature is stronger than previous approaches for obtaining a personal signature from fMRI neurofeedback and provides an indication that a person's learning pattern may be used as a diagnostic tool. Our code has been made available, \footnote{Our code is available via \url{https://github.com/MICCAI22/fmri_nf}} and data would be shared, subject to ethical approvals.
 \end{abstract}

\section{Introduction}

An individual’s ability to learn to perform a specific task in a specific context is influenced by their transient task demand and context-specific mental capacity~\cite{crump2008context}, as well as their motivation~\cite{utman1997performance}, all of which vary considerably between individuals~\cite{ackerman1988determinants}. We hypothesize that the learning pattern is highly indicative of both personal information, such as age and previous experience in performing similar tasks, and personality and clinical traits, such as emotion expressivity, anxiety levels, and even specific psychiatric indications.

Neurofeedback (NF), a closed-loop self-neuromodulation learning procedure, provides a convenient environment for testing our hypothesis. This is because the learning task is well-defined, yet individualized, is presented in a controlled and repeatable manner, and the level of success is measured on a continuous scale of designated neural changes. NF is a reinforcement learning procedure guided by feedback presented depending on self-acquired association between mental and neural states. Mental states that happened to be associated with on-line modulation in the neural target (e.g. lower or higher activity) are rewarded, and eventually result in a desired modification of the brain signal  (i.e. learning success)~\cite{sitaram2017closed,taschereau2022real}. We hypothesize that the established association between internally-generated mental process and neural signal modulation closely signifies personal brain-mind relation and could therefore serve as an informative marker for personality and/or psychopathology. 

fMRI-based neurofeedback enables precise modulation of specific brain regions in real time, leading to sustained neural and behavioral changes. However, the utilization of this method in clinical practice is limited due to its high cost and limited availability, which also hinder further research into the sustained benefit~\cite{sulzer2013real,lubianiker2019process}. In the NF task we consider, one learns to reduce the activity of the Amygdala, while observing a signal that is directly correlated with it. We consider the activity of the rest of the brain as the context or states in which the learning task takes place, and discretize this space by performing clustering. 

A personal signature is constructed by measuring the progress of performing the NF task in each of these clusters. Progress is obtained by comparing the Amygdala activity at a first training session with that observed at the second session for the most similar brain state. Specifically, we consider the difference between the second activity and the one predicted by a neural network that is conditioned on the brain state in the first.

The representation obtained by aggregating these differences across the brain-state clusters is shown to be highly predictive of multiple psychiatric traits and conditions in three datasets: (i) individuals suffering from PTSD, (ii) individuals diagnosed with Fibromyalgia, and (iii) a control dataset of healthy individuals. This predictive power is demonstrated with linear classifiers, in order to demonstrate that the personal information is encoded in an explicit way and to reduce the risk of overfitting by repeating the test with multiple hyperparameters~\cite{asano2019critical}.

\section{Related work}
fMRI is widely used in the study of psychiatric disorders~\cite{calhoun2014chronnectome,oksuz2019magnetic}. Recent applications of deep learning methods mostly focus on fully-supervised binary classification of psychopathology-diagnosed versus healthy subjects in resting state~\cite{dvornek2017identifying,yan2019discriminating}, i.e., when not performing a task. Contributions that perform such diagnosis while performing a task, e.g.,~\cite{bleich2014machine,jacob2019reappraisal,hendler2018social,raz2016psychophysiological,lerner2018abnormal}, focus on comparing entire segments that correspond to phases of the task, and have shown improved ability to predict subjects' traits, w.r.t to resting-state fMRI \cite{gal2022predicting}. Our analysis is based on aggregating statistics across individual time points along the acquired fMRI.

The work closest related to ours applies self-supervised learning to the same fMRI NF data in order to diagnose participants suffering from various psychopathologies and healthy controls~\cite{osin2020learning}. There are major differences in the approaches. First, while our method is based on a meaningful signature (it accumulates meaningful statistics) that indicates learning patterns, their work is based on an implicit embedding obtained by training a deep neural network. Second, while we focus on modeling the success in performing the task over a training period, their method is based on the self-supervised task of next frame prediction, which involves both the preparation (``passive'') and training (``active'') periods (they require more data). Third,  while our method compares progress between two active NF sessions, their method is based on mapping a passive session, in which the participant does not try to self-modulate, and the subsequent active NF session. The methods are, therefore, completely different. Finally, in a direct empirical evaluation, our method is shown to outperform~\cite{osin2020learning} by a sizable gap across all datasets and prediction tasks.

\section{Data}
\label{sec: Data}

Real-time blood-oxygen-level-dependent (BOLD) signal was measured from the right Amygdala region during an interactive neurofeedback session performed inside the fMRI. The data fed into the model went through preprocessing steps using the CONN toolbox. The preprocessing steps are detailed in supplementary Fig. \ref{fig: preprocessing diagram}. In specific parts of the task, subjects were instructed to control the speed of an avatar riding a skateboard using only mental strategies (active phase), while in other parts, subjects passively watched the avatar on the screen (passive phase). During the active phase, local changes in the signal were translated into changing speed, displayed  via a speedometer and updated every three seconds. 

The neurofeedback datasets used in this experiment were part of larger intervention experiments applying multiple training sessions outside the fMRI by using an EEG statistical model of the right Amygdala. The subjects went through pre/post fMRI scans with the model region as target in order to test for changes in the ability to self-regulate this area \cite{fruchtman2021amygdala,keynan2019electrical}.

\noindent{\bf fMRI data \quad}
Each subject performed several cycles of the paradigm in a single session, lasting up to one hour, in a similar fashion to common studies in the field~\cite{paret2019current}. Following each active phase, a bar indicating the average speed during the current cycle was presented for six seconds. Each subject performed $M = 2$ cycles of Passive/Active phases, where each passive phase lasted one minute. Each active phase lasted one minute (for healthy controls and PTSD patients) or two minutes (Fibromyalgia patients). Following previous findings, instructions given to the subjects were not specific to the Amygdala, to allow efficient adoption of individual strategies~\cite{marxen2016amygdala}.

The active sessions were comprised of $T$ temporal samples of the BOLD signal, each a 3D box with dimensions $\mathcal{H\text{[voxels]}\times W\text{[voxels]}\times D\text{[voxels]}}$. We distinguish BOLD signals of the Amygdala signals from signals of other brain regions, each with a different spatial resolution, $\mathcal{H_A \times W_A \times D_A}$, and \\$\mathcal{H_R \times W_R \times D_R}$, respectively. 
Partitioning was done using a pre-calculated binary Region-of-Interest matrix. 

ROI for Rest-of-Brain covers the entire gray matter of the right hemisphere, excluding the right amygdala. This mask was generated with an SPM based segmentation of the MNI brain template. This region was used for providing the feedback during the real-time fMRI experiments.
A visualisation of the selected brain regions, is shown in supplementary Fig.~\ref{fig: ROI visualization}. The Amygdala region of interest was defined in SPM as a 6mm sphere located at MNI coordinates \\ $[21, -1, -22]$. This region was used for providing the feedback during the real-time fMRI experiments. The matrix contains a positive value for voxels that are part of the Amygdala, and a negative value for all others.
Our data is, therefore, comprised of per-subject tuples of tensors: $(\mathbb{R}^{T \times \mathcal{H_A\times W_A \times D_A}},\mathbb{R}^{T \times \mathcal{H_R \times W_R \times D_R}})$. 

In our setting, $T=18$, and we use three datasets in our experiments: (i) \textbf{PTSD}- 51 subjects, (ii) \textbf{Fibromyalgia}- 24 subjects, and (iii) \textbf{Healthy Control}- 87 subjects. The Amygdala parameters and rest-of-brain parameters are through datasets: $(\mathcal{H_A}, \mathcal{W_A},\mathcal{D_A})=(6,5,6)$ and $(\mathcal{H_R}, \mathcal{W_R}, \mathcal{D_R})=(91, 109, 91)$, respectively. Further information on the data acquisition scheme is provided in supplementary Fig. \ref{fig: preprocessing diagram}.

\noindent{\bf Clinical data \quad}
Clinical information about each subject $s$, denoted as $y_s\in\mathbb{R}^l$ was available in addition to the fMRI sequences, consisting of the following information: (1) \textbf{Toronto Alexithymia Scale (TAS-20)}, which is a self-report questionnaire measuring difficulties in expressing and identifying emotions~\cite{bagby1994twenty}, (2) \textbf{State-Trait Anxiety Inventory (STAI)}, which is measured using a validated 20-item inventory~\cite{spielberger1983state}, and (3) \textbf{Clinician Administered PTSD Scale (CAPS-5) 1}, which is the outcome of a clinical assessment by a trained psychologist based on this widely-used scale for PTSD diagnosis~\cite{weathers2013clinician}. For the healthy controls, the following demographic information was also available: (1) {\bf Age} and (2) {\bf Past experience in neuro-feedback tasks}, presented as a binary label (i.e, experienced / inexperienced subject).

\begin{figure}[t]
    \centering 
    \includegraphics[width=0.78\textwidth]{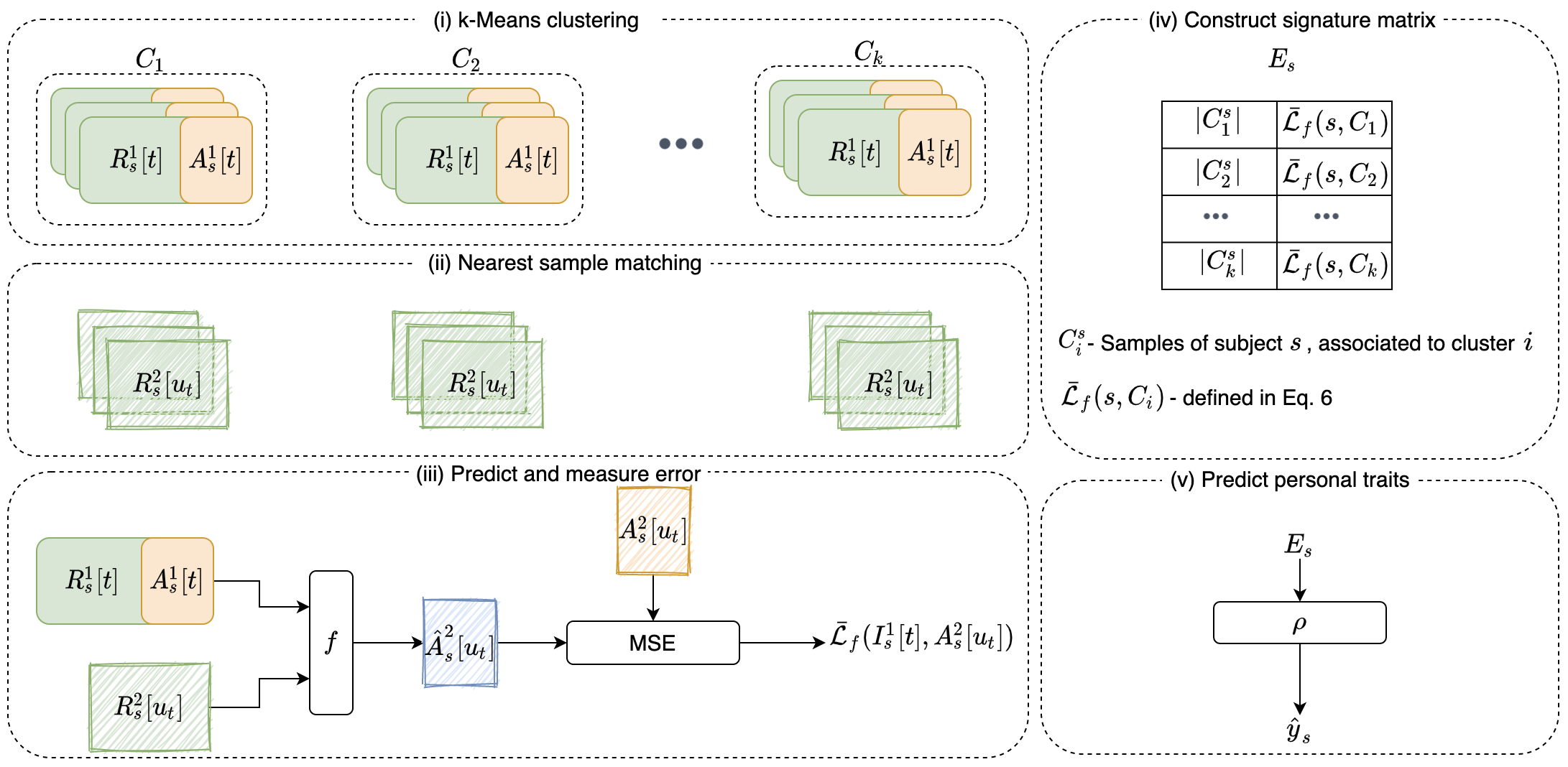}
    \caption{The training steps: (i) Clustering the frames from the first session, based on the regions outside the Amygdala; (ii) Matching each frame from the previous step to the most similar frame from the second session; (iii) Learning to predict the Amygdala activity in the matched frame of the second session, based on the information from the matching frame in the first session and the brain activity outside the Amygdala in the second session; (iv) constructing a signature based on the prevalence of each cluster and the error of prediction for the frames of each cluster; and (v) a linear classifier on top of the obtained signature.}
    \label{fig:overview}
\end{figure}

\section{Method}

Our method aims at obtaining subjects' learning patterns across different ``brain states'' in order to predict their demographic and psychiatric traits. Identification of said brain states was done by clustering all the fMRI frames. The error in predicting the amygdala in each of the states is used as a unique signature of the subjects' learning patterns.

Our network receives processed fMRI samples as inputs, and uses them to predict subjects' demographic and psychiatric criteria. We consider two types of fMRI signals: Amygdala, and rest-of-brain, and train our networks in five steps, as depicted in Fig.~\ref{fig:overview}: \textbf{(i)} identification of prototypical brain states, $C$, by applying a k-means clustering scheme to the fMRI signals; \textbf{(ii)} consider the rest of the brain regions, and identify for each fMRI frame from the first session the most similar frame from the second session; \textbf{(iii)} given a subject's complete brain state {(Amygdala and greater-brain)} at each time-step from the first session, we train a neural network, $f$, to predict the subject's Amygdala state in the matched closest frame from the second session;\textbf{(iv)} create a subject signature by aggregating the prediction error of $f$ in each of the $k$ prototypes; \textbf{(v)} train a linear regression network, $\rho$, to predict subjects' criteria, based on the obtained signature.

\noindent{\bf Data structure } 
For every subject $s\in S$, the dataset contains a series of $M=2$ active sessions, each with $T=18$ samples, denoted as $I_s = \big\{I^m_s[t]\big\}_{t=1, m=1}^{T, M}$. We treat each sample as a pair: (i) an Amygdala sample, $A^m_s[t]$, which is cropped out of $I^m_s[t]$ using a fixed ROI matrix, as explained in Sec.~\ref{sec: Data}, and (ii) a sample of other brain parts, $R^m_s[t]$. 

\noindent{\bf K-means (step i)}
We learn a set of $k$ cluster centroids, $C = \{\mu_1,\dots, \mu_k\}$, based on rest-of-brain training samples from the first session, minimizing the within-cluster-sum-of-squares: 
$\underset{C}{\text{argmin}}\sum\limits_{s,t}\min\limits_{\mu\in C}||R^1_s[t] - \mu||^2$. Note that clustering as well as cluster assignment are carried out independently of the Amygdala samples, and only once for the entire training set.
Each cluster represents a different prototypical brain state and the number of clusters is selected such that for most training subjects, no cluster is underutilized. See Sec.~\ref{sec:experiments}.

\noindent{\bf Amygdala state prediction (steps ii and iii)\quad} For every subject $s$ and time step $t\leq T$ in the first session, we identify $u_t$, the time step during the second session in which the most similar sample was taken: 
 $u_t = \underset{t'\leq T}{\text{argmin}} ||R^1_s[t] - R^2_s[t']||$.

\noindent The obtained pairs $\{(t,u_t)\}_{t=1}^T$ indicate tuples of similar samples $\{(I^1_s[t], I^2_s[u_t])\}$, which we use to train a neural-network, $f$, aimed to predict $A^2_s[u_t]$ (see implementation details in supplementary Fig.\ref{fig:nn overivew}):     $\hat{A}_s^2[u_t] = f\big(R_s^1[t], A_s^1[t], R_s^2[u_t]\big)$.

\noindent $f$ is trained independently of the centroids, minimizing the  MSE loss.

\noindent{\bf Building a signature matrix (step iv) and predicting personal traits (step v)\quad} With the group of centroids, $C$, and the Amygdala-state predictor, $f$, we construct a signature matrix, $E_s\in \mathbb{R}^{|C|\times 2}$, for every subject $s \in S$. 
Each row of this signature corresponds to a specific brain state prototype, and the two columns correspond to: (1) the number of samples in a cluster; and (2) the mean prediction error for that cluster which quantifies the distribute of states for every subject and the deviation from the predicted amygdala activation in each state.
\noindent For every subject $s$ and cluster $C_i\in C$, we define $C_i^s$ as the set of the subject's samples associated with the cluster: 
\noindent We then calculate the average prediction loss of $f$ with respect to each $(s, C_i)$ pair:    $\bar{\mathcal{L}}_f(s,C_i)=\frac{1}{|C_i^s|}\cdot \sum_{t\in C_i^s}||f(R^1_s[t],R^2_s[u_t],A^1_s[t]) - A^2_s[u_t]||^2$,
where $| C_i^s |$ indicates the number of visits of subject $s$ in cluster $i$. The signature matrix $E_s$ has rows of the form 
    $E_s[i] = \big[|C_i^s|, \bar{\mathcal{L}}_f(s, C_i)\big]$.

\noindent $E_s$ is then fed into $\rho$, a linear regression network with an objective to predict $y_s$. Since $E_s$ is a matrix, we use a flattened version of it, denoted as $e_s\in\mathbb{R}^{2\cdot |C|}$. We then predict $\hat{y}_s=\rho(e_s) = G^{\top} e_s + b$, where $G$ is a matrix and $b$ is a vector. $G,b$ are learned using the least squared loss over the training set.

Using a low-capacity linear classifier is meant to reduce the effect of overfitting in this step, which is the only one with access to the target prediction labels. The neural network $f$ is trained on a considerably larger dataset, with samples of a much higher dimension (as every fMRI frame from every subject is a sample). Additionally, its task is a self-supervised one, and is therefore less prone to overfitting to the target labels.

\noindent{\bf The inference pipeline\quad} 
Given an unseen subject $r$, we assign each fMRI frame, $t$, in the first session, to a cluster by finding the prototypes in $C$ closest to $R_r^1[t]$. We also match a frame $u_t$ from the second session to each frame $t$ by minimizing $||R^1_r[t] - R^2_r[u_t]||^2$. A signature $e_r$ is then constructed. 
Finally, the linear predictor $\rho$ is applied to this signature. Implementation details are provided in supplementary Fig. \ref{fig:nn overivew}.

\section{Experiments}
\label{sec:experiments}

Each experiment was repeated five times on random splits. We report the mean and SD. The data is partitioned using a cross-validation scheme between train, validation and test sets, each composed of different subjects, with a 60-20-20 split. The same partition holds for all training stages in each dataset. For each dataset (i.e, Healthy, PTSD, Fibromyalgia), we trained a separate network.

\noindent{\bf Setting k\quad }
In order to assure that all $k$ chosen brain-state-prototypes are visited by all subjects, we evaluate the ratio $\Bar{n}_i = \frac{\sum_{s\in S}|C_i^s|}{T\cdot |S|}$ of brain states assigned to each of the clusters, when applying k-means on the training data. We chose the largest $k$ for which the variance between all $\big\{ \Bar{n}_i\big\}_{i=1}^k$ is relatively small. 
 In order to show that the clusters we got hold meaningful information, we performed two experiments: (a) train an LSTM based network, which receives as input the temporal signal of transitions between clusters, which proved to have predictive power w.r.t subjects’ traits (results are shown in Tab.~\ref{tab: results}); and (b) train a similar LSTM network to predict the next cluster brain state, given past visited clusters. The network accurately predicted the cluster for 41\% of the frames, compared to guessing the mean cluster which yields accuracy of 21\%.  
To better understand the prototypical brain states that are obtained through the clustering process, we have performed an anatomical visualization of the cluster centroids, which represent the prototypical brain states found during the NF task for the Healthy subgroup. See supplementary Fig.~\ref{fig:clustercenters}.

\noindent{\bf Amygdala state prediction\quad}  We trained the network $f$ until its validation loss converged for each of the three datasets - healthy, PTSD and Fibromyalgia. The MSE error obtained is presented in supplementary Tab. \ref{tab:nn mse}. The network is quite successful in performing its prediction task, compared to the simple baseline of predicting the network's input,  $A_s^1[t]$.

\noindent{\bf Predicting a subject's psychiatric and demographic criteria\quad}
We test whether our learned representation, trained only with fMRI images, has the ability to predict a series of psychiatric and demographic criteria not directly related to the neurofeedback task. We used our method to predict (i) STAI and (ii) TAS-20 for PTSD, Fibromyalgia and control subjects, (iii) CAPS-5 for PTSD subjects. Demographic information, (iv) age, and (v) past neuroFeedback experience were predicted for the control subjects. 

Our linear regression scheme, applied to the learned signature vectors, is compared to the following baselines, which all receive the fMRI sequence as input, denoted as $x$: (1) \textbf{Mean prediction} simply predicts the mean value of the training set, (2) \textbf{Conditional LSTM-} The Amygdala sections of the passive and active temporal signals are fed to a neural network, which learns a personal representation for each subject. This learned representation is later used to predict the subject's criteria \cite{osin2020learning}. In contrast to our method, the learned representation of the conditional-LSTM also employs the passive ``watch'' data, which our method ignores.
(3) \textbf{CNN-} A convolutional network with architecture identical to our $f$ network, except for two modifications: (a) the input signal to this network is the entire second sample, $I_s^2[u_t]$ (instead of $R_s^2[u_t]$). This way, the network has access to the same signals our proposed method has; and (b) the decoder is replaced with a fully connected layer, which predicts the label.
(4) \textbf{clinical prediction-} an SVM regression with the RBF kernel performed on every trait, according to the other traits.
(5) \textbf{Raw difference-} A similar signature matrix, with the $\bar{\mathcal{L}}_f$ value replaced by the average norm of differences between  amygdala signals for every pair $\{(t, u_t)\}$. Rows of the resulting matrix are: $\tilde{E}_s[i] = \big[|C_i^s|,\frac{1}{|C_i^s|}\cdot \sum\limits_{t\in C_i^s} ||A^1_s[t]-A^2_s[u_t]||^2\big]$; and (6, 7) \textbf{partial $E_s$-} A network trained using only one of  $|C_i^s|$ or $\bar{\mathcal{L}}_f$; (8) \textbf{ClusterLSTM-} an LSTM based network, which receives as input the sequence of cluster memberships per frame, and predicts the subject's traits according to it. To show that the neurofeedback learning that occurs across sessions is what is important, rather than the expected state of amygdala based on "Rest-of-Brain" state, we implemented (9) \textbf{No Feedback-} which predicts the Amygdala state given “Rest-of-Brain” state for the sample, without pairing samples from different sessions. 
(10) \textbf{Alternative clustering-} In step (i), the clustering objective is changed, such that it depends on the subjects' complete brain state:      $\underset{C}{\text{argmin}}\sum\limits_{s,t}\min\limits_{\mu\in C}||I^1_s[t] - \mu||^2$. Lastly, to demonstrate the importance of using the Amygdala itself, we run baseline (11) \textbf{Alternative ROI-} a framework identical to ours, but with neural area of focus shifted from the Amygdala to the primary motor cortex, an area of dimensions $\mathcal{H=W=D}=8$, which is presumed not to take part in the performance of the NF task.

\begin{table}[t]
	\begin{minipage}{0.69\linewidth}

\resizebox{\linewidth}{!}{
\begin{tabular}{@{}l@{~}c@{~~}c@{~~}c@{~~}c@{~~}c@{~~}c@{~~}c@{~~}c@{~~}c@{~~}c@{}}
%\toprule
  & \multicolumn{3}{c}{Healthy} &  & \multicolumn{2}{c}{Fibromyalgia} &  & \multicolumn{3}{c}{PTSD} \\ %\cmidrule{2-4} \cmidrule{6-7} \cmidrule{9-11} 
  & Age$\downarrow$  & TAS$\downarrow$    & STAI$\downarrow$ &  & TAS$\downarrow$        & STAI$\downarrow$      &  & TAS$\downarrow$    & STAI$\downarrow$   & CAPS-5$\downarrow$ \\ \hline
    Mean & $13.7 \pm 1$ & $121.7 \pm 6$ & $79.3 \pm 6$ & & $98.6 \pm 7$ & $78.5 \pm 5$ & & $153.0 \pm 11$ & $159.5 \pm 13$ & $119.3 \pm 10$\\
  \cite{osin2020learning} & $10.1\pm 1$  & $81.6 \pm 7$ & $67.4\pm 6$ &  & $44.0 \pm 5$     & $73.1 \pm 3$   &  & $99.2\pm 5$ & $132.7\pm 8$ & $85.4 \pm 8$ \\
CNN & $17.2 \pm 1$  & $110.3 \pm 12$ & $81.6 \pm 7$ & & $65.2 \pm 10$ & $90.1 \pm 9$ &  & $105.0 \pm 11$ & $166.3 \pm 33$ & $98.4 \pm 16$ \\
SVM & $13.0 \pm 2$  & $100.0 \pm 12$ & $78.2 \pm 8$     &  & $74.3 \pm 7$     & $77.9 \pm 11$    &  & $148.3 \pm 12$ & $150.0\pm 11$ & $117.4 \pm 14$ \\
%\cmidrule{1-11}
Ours & $\textbf{9.3} \pm 1$  & $\textbf{79.0} \pm 3$ & $\textbf{55.2} \pm 7$ & & $\textbf{35.3} \pm 8$ & $\textbf{62.3} \pm 7$ &  & $\textbf{89.1} \pm 11$ & $\textbf{97.0} \pm 10$ & $\textbf{84.3} \pm 10$\\
%\cmidrule{1-11}
Ablation $\Tilde{E}_s$ & $13.2 \pm 1$ & $98.4\pm 15$ & $73.2 \pm 9$ &  & $77.8 \pm 7$ & $77.4 \pm 11$ &  & $102.2 \pm 20$ & $194.8 \pm 6$ & $136.0 \pm 14$ \\
Ablation $\mathcal{L}_f$ & $10.3\pm 1$ & $91.6 \pm 4$ & $70.0\pm 9$     &  & $84.4 \pm 13$     & $81.9 \pm 14$    &  & $119.6\pm 11$ & $141.0\pm 9$ & $119.2 \pm 19$ \\
Ablation $|C_i|$ & $11.0\pm 2$ & $84.2\pm 9$ &    $68.1\pm 9$  &  & $113.6 \pm 32$     & $109.4\pm 20$    &  & $137.5\pm 18$ & $137.8 \pm 20$ & $151.5\pm 19$ \\
ClusterLSTM & $12.1 \pm 1$ & $96.8 \pm 6$ & $68.8 \pm 8$ & & $75.7 \pm 9$ & $90.0 \pm 6$ & & $118.1 \pm 12$ & $155.0 \pm 14$ & $108.0 \pm 11$\\
No Feedback  & $13.7 \pm 2$ & $125.0 \pm 9$ &    $75.0 \pm 8$  &  & $90.0 \pm 8$     & $77.7 \pm 7$    &  & $165.3 \pm 16$ & $167.6 \pm 18$ & $139.0 \pm 11$ \\
Alt. clustering  & $12.3\pm 1$ & $114\pm 5$ &    $72.5\pm 9$  &  & $70.2 \pm 8$     & $80.3\pm 8$    &  & $98.3\pm 12$ & $174.0 \pm 20$ & $103.6\pm 12$ \\
Alt. ROI & $13.5\pm 1$ & $125.5\pm 7$ &    $79.0\pm 10$  &  & $96.3 \pm 9$     & $84.5\pm 8$    &  & $135.6\pm 16$ & $135.8 \pm 19$ & $128.7\pm 5$ \\
  
\hline
\end{tabular}
}
        \caption{Traits prediction results (MSE)}
        \label{tab: results}
    \end{minipage}\hfill
    \begin{minipage}{0.31\linewidth}
	    \centering
    	\includegraphics[width=\textwidth]{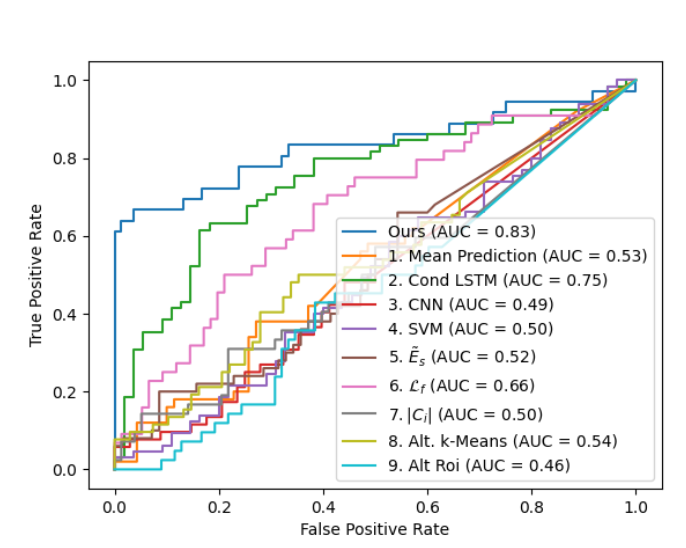}
    	\captionof{figure}{Experience ROC}
    	\label{fig: ROC}
    \end{minipage}
    \vspace{-.5cm}
\end{table}

The full results are shown in Tab.~\ref{tab: results} for performing regression on age, TAS, STAI and CAPS-5. Our results are statistically significant, and the $p$-values of the corrected re-sampled t-test between our method and the baseline methods is always lower than $0.01$. As can be seen, the baseline of~\cite{osin2020learning} greatly outperforms the mean prediction and both the CNN classifier and the one based on the clinical data. However, it is evident that across all three datasets our method outperforms this method, as well as all ablations, by a very significant margin in predicting the correct values for both demographic and psychiatric traits. 

Fig.~\ref{fig: ROC} presents classification results for past-experience information, which is only available for the healthy control subjects. Here, too, our method outperforms the baselines and ablation methods. Specifically, it obtains an AUC of $0.83 \pm 0.03$, while the method presented by ~\cite{osin2020learning} obtains an AUC of $0.75 \pm 0.03$. 

The ablation experiments provide insights regarding the importance of the various components. First, modeling based on an irrelevant brain region, instead of the Amygdala, leads to results that are sometimes worse than a mean prediction. Similarly, predicting using raw differences in the Amygdala activity (without performing prediction), is not effective. It is also important to remove the Amygdala from the clustering procedure, keeping this region and outside regions separate. The variant based on the prediction error alone seems to be more informative than that based only on cluster frequency. However, only together do they outperform the strong baseline of~\cite{osin2020learning}.

\begin{figure}
    \centering
    \begin{tabular}{ccc}
    \includegraphics[clip,width=0.33\linewidth]{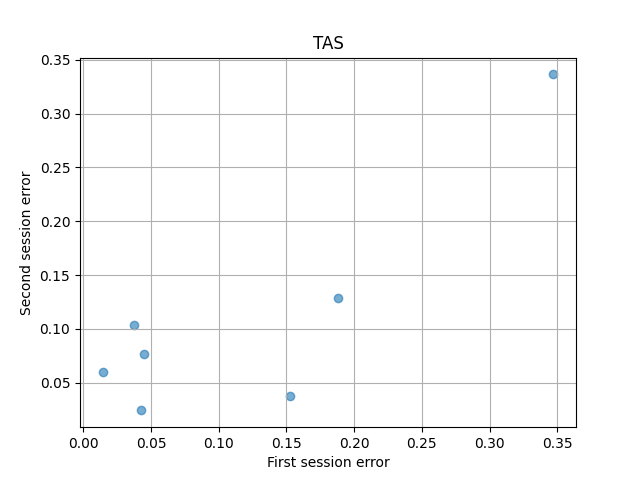}&
    \includegraphics[clip,width=0.33\linewidth]{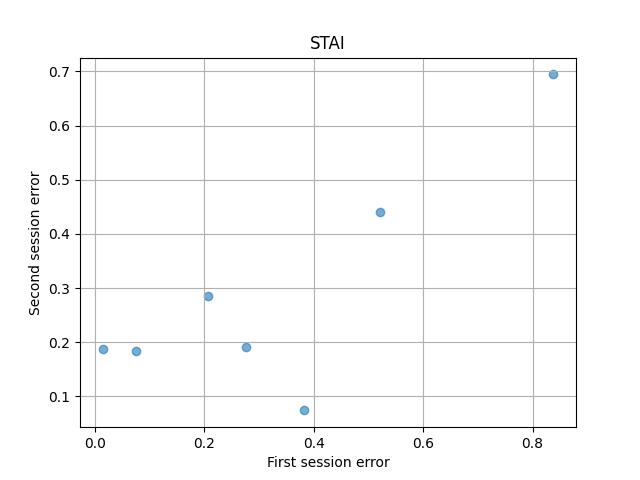}&
    \includegraphics[clip,width=0.33\linewidth]{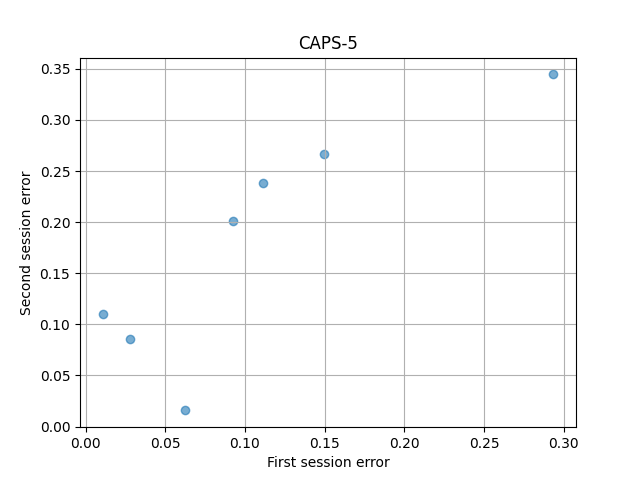}\\
    (a) & (b) & (c)\\
    \end{tabular}
    \caption{The normalized error in the second session  as a function of that of the first session. Each point reflects a single patient of the PTSD cohort. (a) TAS, (b) STAI, (c) CAPS-5.  The normalized error is computed as the absolute error divided by the ground truth value.}
    \label{fig:longitudinal}
\end{figure}

\noindent{\bf Longitudinal data\quad} In order to show the robustness of our method over time, we performed inference of our network (which was trained using data from the first session of each subject) on fMRI scans that were obtained a few months later. As can be seen from Fig.~\ref{fig:longitudinal}, while there are not many subjects that have both sessions, there seems to be a correlation between the errors in the first session and the errors in the second one. More data is requires in order to further study the validity of the model beyond the first collection point.

\section{Conclusion}

NF data offers unique access to individual learning patterns. By aggregating the deviation between actual and predicted learning success across clusters of brain activities, we obtain a signature that is highly predictive of the history of a person, as well as of their clinical test scores and psychiatric diagnosis.  

The presented method provides a sizable improvement in performance over previous work. Perhaps even more importantly, the obtained signature is based on explicit measurements that link brain states to the difference between actual and expected learning success, while previous work was based on an implicit embedding that is a by-product of training a network to predict a loosely related task of predicting transient signal dynamics. By accurately predicting the psychiatric tests score, our method provides a step toward an objective way to diagnose various clinical conditions.

\section*{Acknowledgments} 
This project has received funding from the European Research Council (ERC) under the European Union’s Horizon 2020 research and innovation programme (grant ERC CoG 725974), and the ISRAEL SCIENCE FOUNDATION (grant No. 2923/20) within the Israel Precision Medicine Partnership program.

%\newpage
\bibliographystyle{splncs04}

\newpage
\pagenumbering{gobble}
\appendix
\renewcommand\thefigure{\thesection.\Roman{figure}}    
\setcounter{figure}{0}    

\renewcommand\thetable{\thesection.\Roman{table}}    
\setcounter{table}{0}

\begin{figure}[t]
    \centering
    \includegraphics[width=\textwidth]{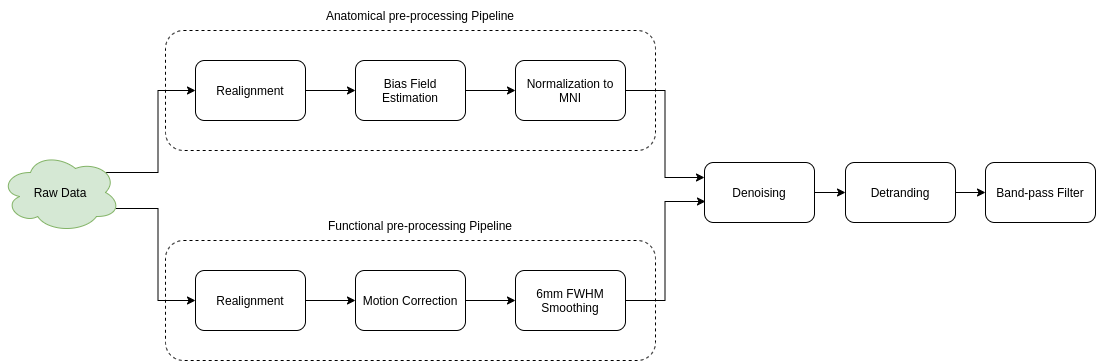}
    \caption{Structural and functional scans were obtained with a 3.0T Siemens PRISMA MRI system.
    The CONN MATLAB toolbox~\cite{whitfield2012conn} was used for functional volumes realignment, motion correction, normalization to MNI space and spatial smoothing with an isotropic 6-mm FWHM Gaussian kernel. Subsequently de-noising and de-trending regression algorithms were applied, followed by bandpass filtering in the range of 0.008-0.09 Hz. The frequencies in the bandpass filter reflect the goal of modeling the individual throughout the session, while removing the effects of the fast paced events that occur during the neurofeedback session. This filtering follows previous work~\cite{osin2020learning} for a fair comparison.}
        \label{fig: preprocessing diagram}
\end{figure}

\begin{figure}[b]
    \centering 
    \includegraphics[width=\textwidth]{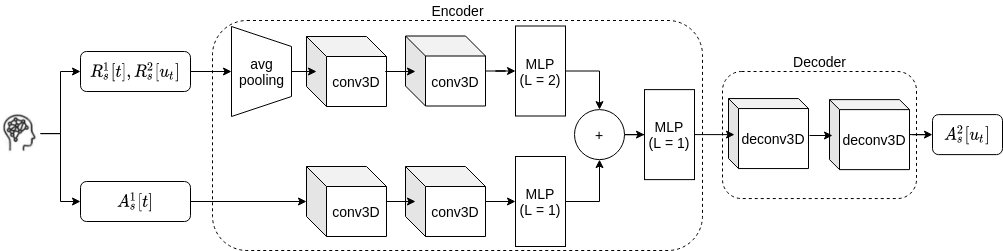}
    \caption{Architecture of $f$, our Amygdala prediction Neural Network. We trained the network $f$, until convergence of its validation loss for each of the three datasets. The hyper-parameters of the network $f$ were selected according to a grid search using the cross validation scores on the validation set. For training, we used an Adam optimizer, with initial learning rate of 0.001, and a batch sizes of 16.}
    \label{fig:nn overivew}
\end{figure}

\begin{figure}[t]
    \centering
    \begin{subfigure}[b]{0.37\textwidth}
            \includegraphics[width=\linewidth]{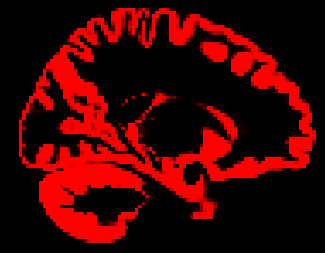}    
    \end{subfigure}
        \begin{subfigure}[b]{0.18\textwidth}
            \label{fig:aapic2}% label for this sub-figure
            \includegraphics[width=\textwidth]{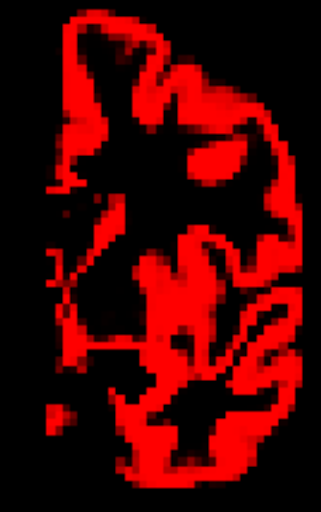}
        \end{subfigure}
    \caption{(a) Mid-sagittal slice of Rest of brain mask; and (b) a coronal slice of it.}
        \label{fig: ROI visualization}
\end{figure}

\begin{table}[b]
\caption{The MSE error of network $f$ in comparison to the simple baselines of predicting the activations in the network's input $A_s^1[t]$ and predicting the mean activation of each Amygdala's voxels}
\label{trainf}
\centering
\begin{tabular}{lccc}
%\toprule
Method                          & Healthy & Fibromyalgia & PTSD  \\
\hline
Mean Prediction           & $0.1151 \pm 3\cdot 10 ^{-3}$        &  $0.0780 \pm 1\cdot 10 ^{-3}$  &  $0.0770 \pm 1\cdot 10 ^{-3}$   \\
Predicting $A_s^1[t]$     &  $0.1087 \pm 2\cdot 10 ^{-3}$       & $0.0833 \pm 4\cdot 10 ^{-3}$  &  $0.0781 \pm 3\cdot 10 ^{-3}$   \\
$f(I_s^1[t], R_s^2[u_t])$ & $\textbf{0.0735} \pm 2\cdot 10 ^{-3}$         &  $\textbf{0.0561} \pm 2\cdot 10 ^{-3}$ &  $\textbf{0.0550} \pm 1\cdot 10 ^{-3}$  \\
%\bottomrule
\end{tabular}
\label{tab:nn mse}
\end{table}

\begin{figure}
    \centering
    \begin{tabular}{cc}
    \includegraphics[trim={0cm 0 0 3cm},clip,width=0.4\linewidth]{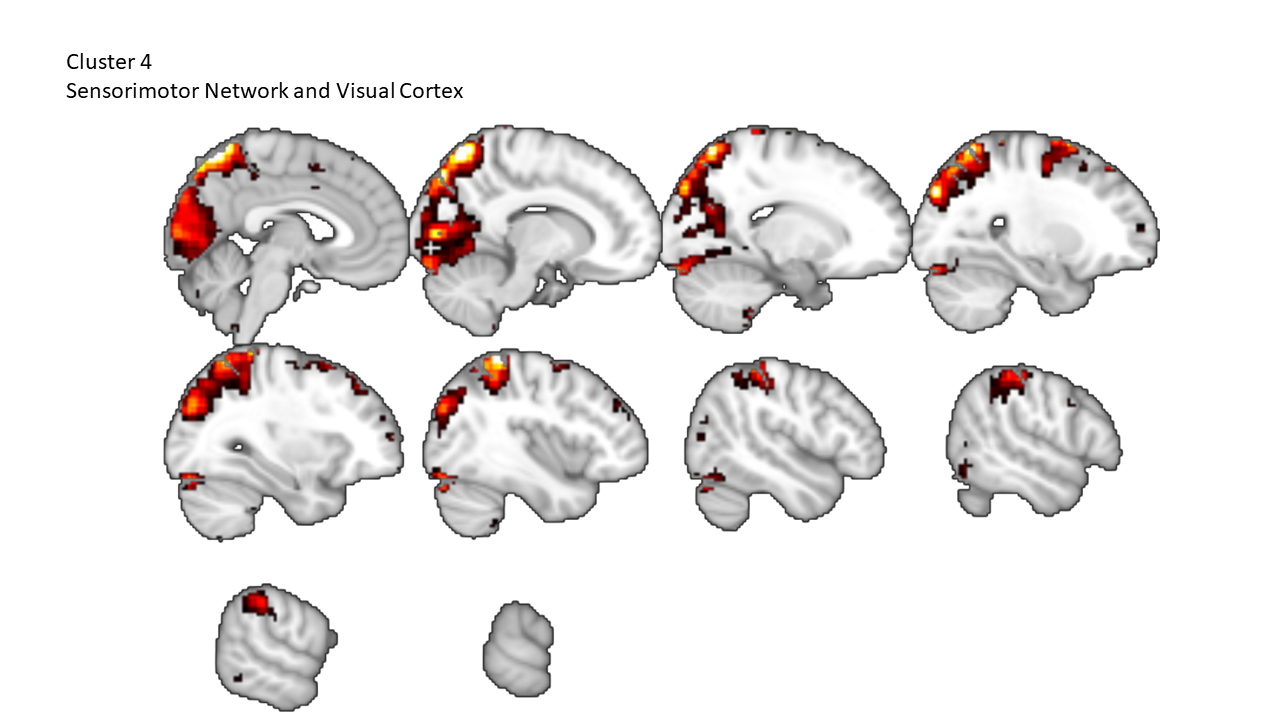}&
    \includegraphics[trim={0cm 0 0 3cm},clip,width=0.4\linewidth]{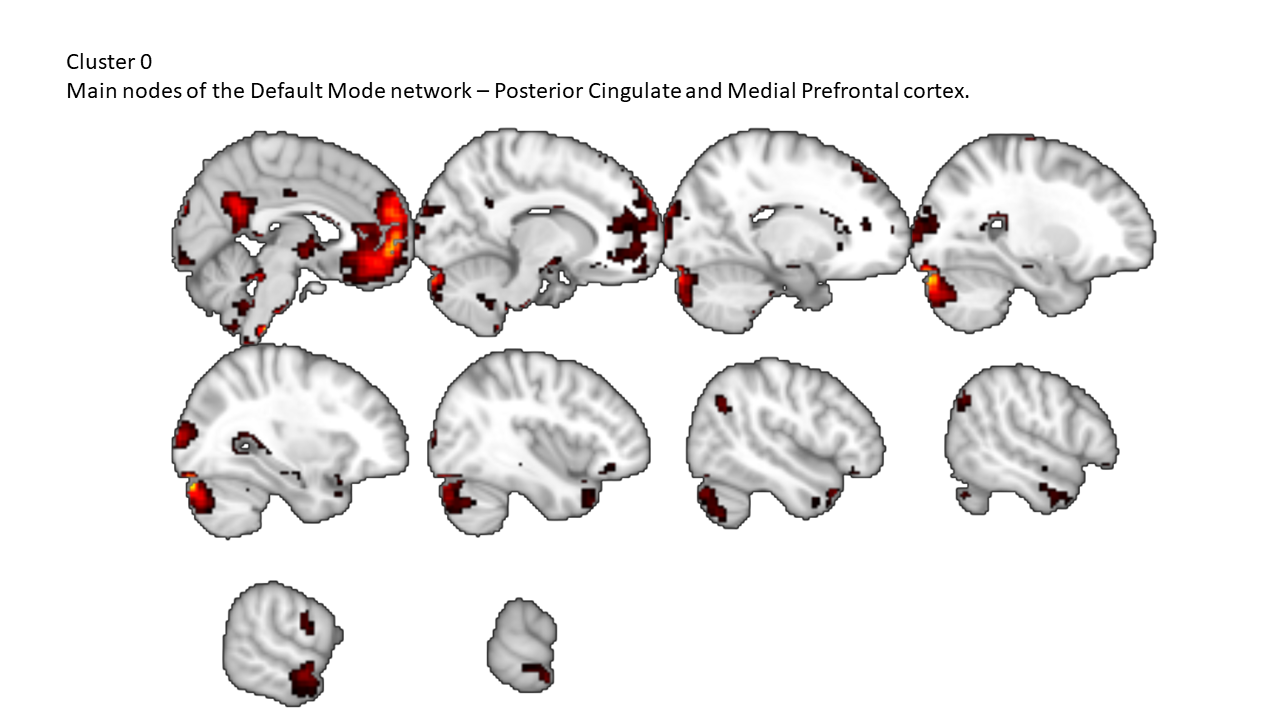}\\
    (a) & (b)\\
    \includegraphics[trim={0cm 0 0 3cm},clip,width=0.4\linewidth]{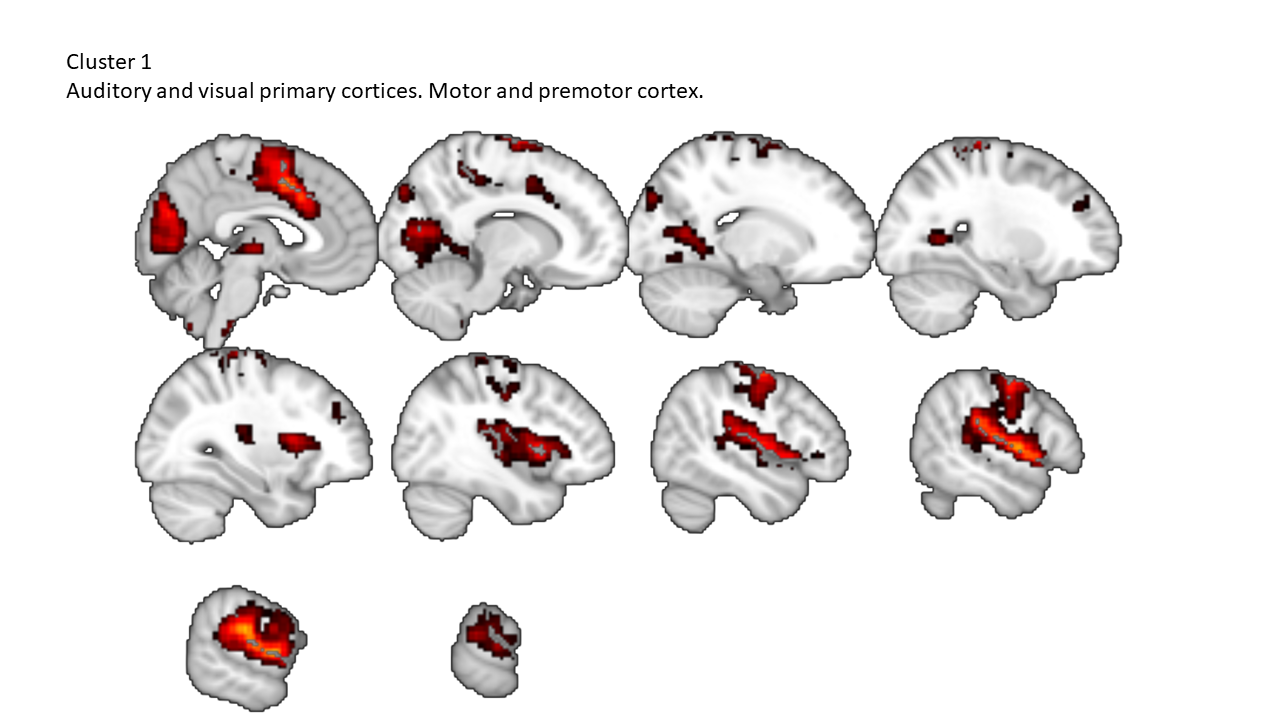}&
    \includegraphics[trim={0cm 0 0 3cm},clip,width=0.4\linewidth]{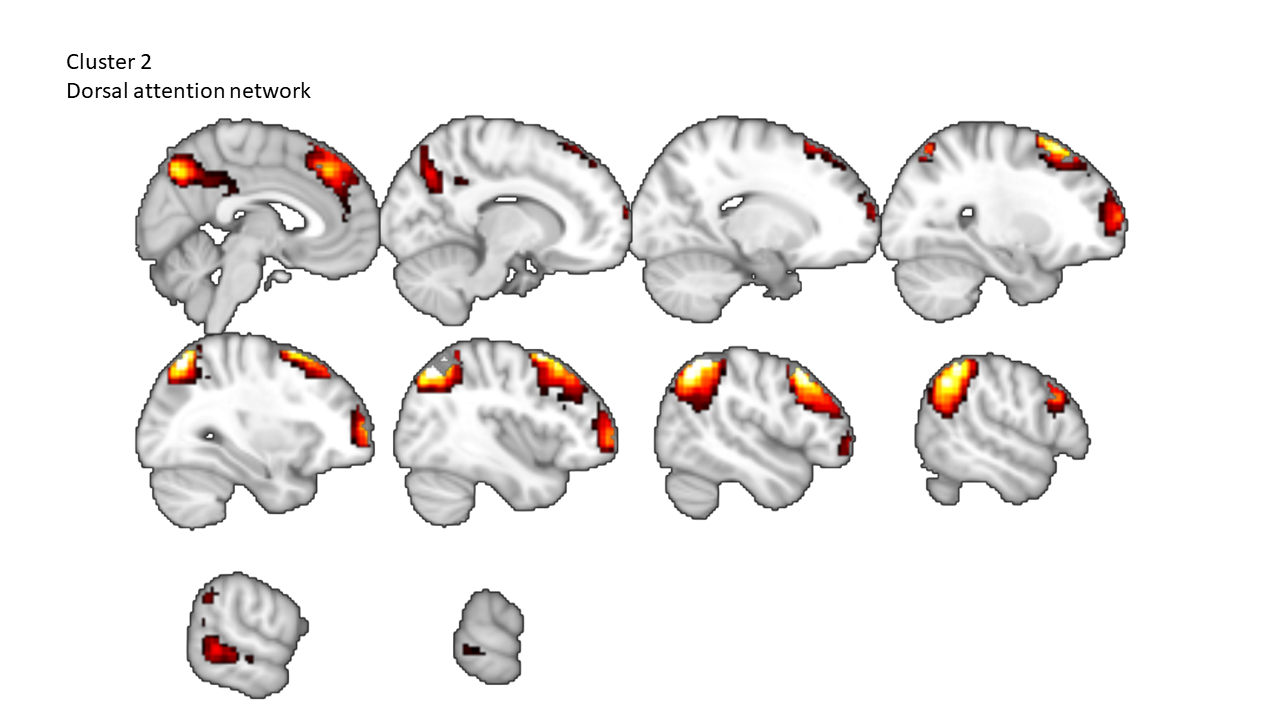}\\
    (c) & (d)\\
    \includegraphics[trim={0cm 0 0 3cm},clip,width=0.4\linewidth]{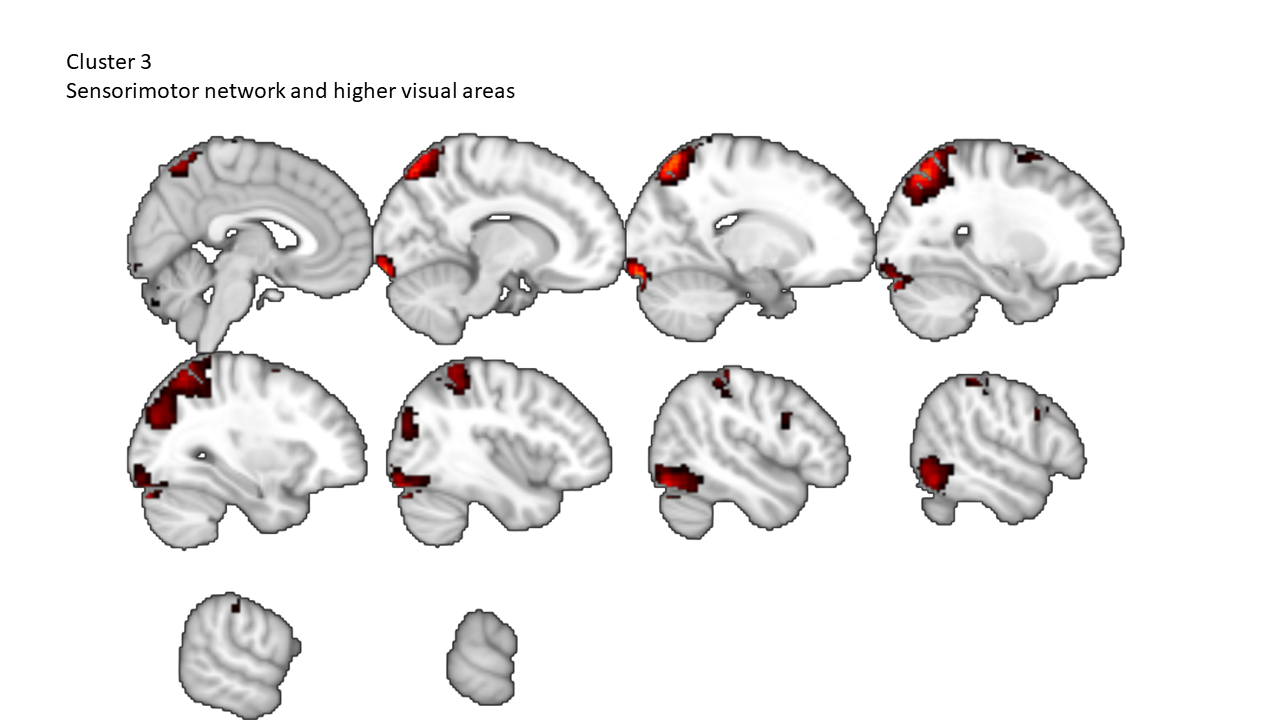}& \\
    (e)\\
    \end{tabular}
    \caption{Visualization of the cluster centroids which represent the prototypical brain states found during the NF task for the Healthy subgroup. The resulting maps show the main activated nodes in each state. Clusters are distinct in their spatial arrangement, which supports the relevance of using clustering for this purpose. 
    In this figure, the five prototypical clusters obtained on the healthy individuals dataset are presented. (a) Sensorimotor Network and Visual Cortex. (b) Main nodes of the Default Mode network (c) Main nodes of the Salience Network. (d) Dorsal attention network. (e) High visual areas (Ventral and Dorsal Stream).}
    \label{fig:clustercenters}
\end{figure}

\end{document}